\documentclass[10pt,journal]{IEEEtran}
\usepackage{epsfig} 	
\usepackage{siunitx}	
\usepackage[makeroom]{cancel}
\usepackage{amsmath, amssymb, amsfonts}		
\usepackage{bm}
\newcommand{\norm}[1]{\left\lVert#1\right\rVert} 	
\usepackage{relsize}						
	
\usepackage{booktabs}						
\usepackage{array}							
\usepackage{multirow}						
\usepackage{microtype}
\usepackage[font=small,skip=0pt]{caption}	

\usepackage{pgfplots}
\pgfplotsset{width = 10cm, compat = 1.5}
\usepgfplotslibrary{groupplots}
\usepgfplotslibrary{fillbetween}
\usetikzlibrary{patterns}			

\begin{document}

\title{Minimizing Energy Consumption and Peak Power of Series Elastic Actuators: a Convex Optimization Framework for Elastic Element Design}

\author{Edgar Bol\'ivar, ˜\IEEEmembership{Student Member, ˜IEEE}, Siavash Rezazadeh, ˜\IEEEmembership{Member, ˜IEEE}, and Robert Gregg, ˜\IEEEmembership{Senior Member, ˜IEEE}
	\thanks{
	This work was supported by the National Science Foundation under Award Number 1830360 and the National Institute of Child Health \& Human Development of the NIH under Award Number DP2HD080349. The content is solely the responsibility of the authors and does not necessarily represent the official views of the NIH or the NSF. R. D. Gregg holds a Career Award at the Scientific Interface from the Burroughs Wellcome Fund. E. Bol\'ivar, S. Rezazadeh, and R. D. Gregg are with the Departments of Bioengineering and Mechanical Engineering, The University of Texas at Dallas, Richardson, TX, 75080, USA. Email: \{ebolivar, rgregg\}@ieee.org, siavash.rezazadeh@utdallas.edu.
	}
}

\maketitle    

\begin{abstract}
Compared to rigid actuators, Series Elastic Actuators (SEAs) offer a potential reduction of motor energy consumption and peak power, though these benefits are highly dependent on the design of the torque-elongation profile of the elastic element. In the case of linear springs, natural dynamics is a traditional method for this design, but it has two major limitations: arbitrary load trajectories are difficult or impossible to analyze and it does not consider actuator constraints. Parametric optimization is also a popular design method that addresses these limitations, but solutions are only optimal within the space of the parameters. To overcome these limitations, we propose a non-parametric convex optimization program for the design of the nonlinear elastic element that minimizes energy consumption and peak power for an arbitrary periodic reference trajectory. To obtain convexity, we introduce a convex approximation to the expression of peak power; energy consumption is shown to be convex without approximation. The combination of peak power and energy consumption in the cost function leads to a multiobjective convex optimization framework that comprises the main contribution of this paper. As a case study, we recover the elongation-torque profile of a cubic spring, given its natural oscillation as the reference load. We then design nonlinear SEAs for an ankle prosthesis that minimize energy consumption and peak power for different trajectories and extend the range of achievable tasks when subject to actuator constraints.
\end{abstract}

\section{Introduction}
\IEEEPARstart{H}{uman-robot} interaction represents one of the biggest challenges of this decade for the software and hardware design of robotic systems. From the hardware perspective, elasticity represents a mechanical principle that empowers the current generation of human-friendly robotic systems. In general, elasticity can be considered in multiple components of the robot; one popular strategy considers an elastic element connected in series between traditional rigid actuators and the load. This concept, pionered by Pratt and Williamson \cite{Pratt1995}, is known as series elastic actuators (SEAs). In general, the rigid actuator could be an electric motor or a hydraulic or pneumatic cylinder; in this work, the analysis considers only electric motors. Powered prosthetic legs \cite{Hollander2006,Rouse2014,Bolivar2016}, as well as humanoid \cite{Hurst2008} and manufacturing robots working in close contact with human users \cite{bischoff2010kuka}, are important applications of this technology. In traditional SEAs, the motor is connected to a high-ratio linear transmission, then an elastic element connects the transmission's output to the load \cite{Paine2014}. Designs with a low ratio transmission are less common, but still possible due to the increasing supply of high torque motors \cite{Hirzinger2002, Schutz2016}.


The architecture of SEA's offers important benefits to the actuation of robotic systems. The elastic element in an SEA decouples the reflected inertia of the rigid actuator and the inertia of the load \cite{Hurst2008}. In addition, the spring can store elastic energy and release it with enormous power. SEAs also work as a soft load cell, suitable for measuring and controlling force generation \cite{Robinson1999}. Robots using SEAs exploit these important characteristics in order to reduce the energy lost during impacts \cite{Hurst2008}, improve the safety of the human and robots \cite{Bicchi2004}, move loads with higher velocities \cite{Braun2013}, reduce energy consumption of the system \cite{Vanderborght2006,Jafari2013, Ding2017}, and decrease peak motor power \cite{Jafari2011,Jafari2013, Hollander2006}, so a smaller/lighter motor can be used. All these benefits are subject to the design of the SEA's elastic element and the motion task.

Traditionally, the design of elastic elements for SEAs to minimize energy consumption and/or peak power follows two main approaches: natural dynamics \cite{Verstraten2016} or parameterized optimization \cite{Hollander2006}. Natural dynamics associates first principles with the benefits of elastic elements, which provides intuition for the designer. However, this formulation is limited by the kind of trajectories that can be analyzed. For instance, the desired motion of the load may not correspond to the natural frequency of a conservative elastic element. Even if the required motion matches a natural frequency of oscillation, holding the motor's initial position may require (depending on the transmission and the load) a reactionary torque which dissipates energy by Joule heating (i.e., copper losses). 
For this configuration, a more accurate formulation considers the natural frequency of the double-mass single-spring system formed by the elastic element, the motor, and load inertia \cite{Bolivar2017}. Note that the natural frequency of this system may not have an analytic solution once nonlinear elastic elements are considered. 
Natural dynamics also do not explain how to design the elastic element when actuator constraints are considered. Maximum deflection of the elastic element, maximum torque, and maximum velocity of the motor are important constraints imposed by the construction of the device; neglecting them may lead to an infeasible design.

The second approach, parameterized optimization, describes the elastic element as a set of parameters, e.g., stiffness constant for the case of linear springs, and optimizes over the set of parameters \cite{Rouse2014}. In contrast to natural dynamics, constraints are explicitly included in the optimization problem. Parameterized optimization, however, has limitations. It only guarantees a minimum value within the scope of the parameter space; optimization results are typically local and sensitive to initial conditions, and rely on gradient-descent methods that assume differentiability of the objective function with respect to the parameters \cite[p.~9]{Boyd2004}. In addition, parameterization assumes a specific shape for the elastic element, which may limit the types of springs that can be considered. For instance, a nonlinear conservative spring could further reduce the energy consumption compared to a parameterized quadratic spring.
The same limitations arise when the objective is to simultaneously minimize peak power and energy; thus the design of elastic elements that reduce both energy consumption and peak power for SEAs remains an open question.

\subsection*{Our Contribution}

In this work, the elastic element of an SEA is defined as a function $f: \mathbb{R} \rightarrow \mathbb{R}$, where
\begin{equation}
\tau_{ela}=f(\delta), \label{eq:EOM_ElasticElementFunction}
\end{equation}
and $\delta$ is the elongation and $\tau_{ela}$ the torque of the elastic element. \textit{Our contribution is to specify $f(\delta)$ as the solution of a convex quadratically constrained quadratic program (convex-QCQP), such that it minimizes energy consumption and peak power of the electric motor while satisfying actuator constraints}. A preliminary version of this work \cite{Bolivar2017} discussed a similar formulation to minimize only energy consumption. In this paper, we extend this formulation to minimize both peak power and energy consumption as a multiobjective optimization problem. Convexity of the optimization problem relies on a convex approximation to the expression of peak power. Convexity is important because it guarantees that the solution of the optimization problem is a global optimum, i.e., if the cost is only energy consumption, the resulting elastic element reduces more of this energy than any other conservative elastic element.

We constrain $f(\delta)$ to be a strictly increasing monotonic function to ensure that it represents a conservative elastic element. In contrast to previous design methods, we do not assume that $f(\delta)$ is defined by a set of parameters or that the reference motion of the load resembles natural dynamics. Our design approach also considers actuator constraints, allowing the design of nonlinear SEAs that can perform tasks that would be infeasible for linear SEAs or rigid actuators. In addition, our methodology considers arbitrary periodic trajectories, which can include a combination of periodic tasks. For example, an SEA designed for the ankle joint of a powered prosthetic leg to minimize energy consumption and peak power during walking and running.

The content is organized as follows. A description of the energy flow, expressions of peak power, and modeling of an SEA are provided in Section~\ref{sec:DescriptionModeling}. This introduces the multiobjective convex optimization problem that supports the design of the elastic element in Section~\ref{sec:ElasticElementDesign}. Section~\ref{sec:Simulation results} considers two different tasks for the application of the optimization framework: the natural oscillation of a nonlinear spring and the design of an SEA for the ankle joint of a powered prosthetic leg. Section~\ref{sec:ConclAndDiscuss} discusses and concludes the proposed methodology and corresponding results.

\section{System Description and Modeling}\label{sec:DescriptionModeling}

SEAs are mechatronic devices that transduce electrical energy into mechanical and vice versa. From the energy perspective, they are similar to traditional electric motors; however, their capability to store and release elastic energy creates an additional opportunity to reduce the energy consumption and peak power of their electric motors. This section describes the energy flow of SEAs as an introduction to our formulation.

\begin{figure}[t]
\begin{center}
\includegraphics[scale=1]{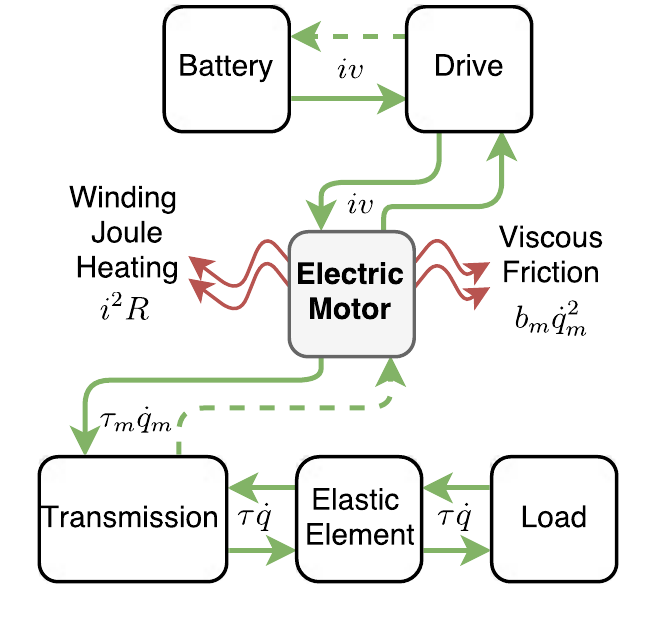}
\end{center}
\caption{Energy flow of an SEA: Dashed lines indicate that the energy path may or may not exist depending on the construction of the device. For instance, energy flowing from the drive to the battery requires drivers capable of regeneration. Energy flowing from the load to the electric motor requires that the load is high enough to backdrive the motor-transmission system.}
\label{fig:EnergyFlowSEA} 
\end{figure}

In this work, we analyze SEAs powered by a battery and an electric motor, a typical scenario for portable devices such as wearable robots \cite{Hurst2008}.
The corresponding energy flow and main components are illustrated in Fig.~\ref{fig:EnergyFlowSEA}. In practice, every component in the system is capable of dissipating energy. For example, the battery self-discharges, the motor drive produces Joule heat, friction in the transmission generates heat, and the elastic elements are not purely elastic (i.e., dissipate energy through their viscous behavior). We concentrate on the energy consumed by the motor since it is the largest consumption in the system. In addition, the energy consumed by other elements of the SEA may be lumped in the expressions of energy of the motor. For instance, viscous friction at the transmission can be considered as additional motor viscous friction.

Energy flow in an electric motor occurs in two principal modes of operation: actuator and generator mode. As an actuator, the electric motor receives electrical energy from the battery/driver and converts it into mechanical energy and heat. Traditional motor drives and mechanical transmissions are designed so that the electric motor can always operate in this mode. A more interesting scenario occurs when the motor works as a generator. For example, when the motor is decelerating a load, the kinetic energy of the load and elastic energy of the spring is transferred to the motor's rotor to store it as electrical energy in the battery.

However, traditional SEAs may not function in generator mode. These designs typically have linear transmissions with high reduction ratios. The reflected inertia of the motor after the transmission, which is proportional to the reduction ratio squared, is normally very high compared to the load. For example, three recent SEA designs reflect output inertia of \SI{360}{\kilo\gram}, \SI{270}{\kilo\gram}, and \SI{294}{\kilo\gram} for the UT-SEA \cite{Paine2014}, Valkyrie's SEA \cite{Zhao2015}, and THOR-SEA \cite{Knabe2014} respectively, as indicated by \cite{Schutz2016}. As a consequence, the system requires a high load to backdrive.  An additional limitation is the motor driver. In order to regenerate energy, motor drivers should be selected such that the electrical energy recovered from the motion of the rotor can flow back to charge the battery \cite{Seok2015}. 

\textit{In this work, we assume the SEA has been designed such that energy can flow from the load to the energy source and vice versa. In other words, the load is high enough to backdrive the motor and adequate electronics allow energy to flow to and from the battery}. In this case, the energy it consumes, $E_m$, is given by
\begin{equation}
E_m = \int_{t_0}^{t_f}  \underbrace{\frac{ \tau_m^2}{k_m^2}}_{\substack{\text{Winding} \\ \text{Joule} \\ \text{heating}}}+\underbrace{\tau_m\dot{q}_m }_{\substack{\text{Rotor} \\ \text{mechanical} \\ \text{power}}}   dt,
\label{eq:EnergyConsumedByMotor_IntegralExpression}
\end{equation}
where $t_0$ and $t_f$ are the initial and final times of the trajectory respectively, $k_m$ is the motor constant, $\tau_m$ the torque produced by the motor, and $\dot{q}_m$ the motor's angular velocity. Notice that energy associated with Joule heating can be also written as $i_m^2R$, since $\tau_m = i_m k_{t}$ and $k_m=k_{t}/\sqrt{R}$, where $i_m$ is the electric current flowing through the motor, $R$ the motor terminal resistance, and $k_{t}$ the motor torque constant \cite{Verstraten2015}. 

 In addition to energy consumption, mechanical peak power of the motor is an important property to analyze because it is related to the weight and size of the actuator. Its mathematical expression is written as follows:
 \begin{equation}
 \text{Peak power} = \norm{\tau_m\dot{q}_m}_\infty.
 \label{eq:PeakPower}
 \end{equation}  
 
\subsection{Modeling}
\begin{figure}[t]
\begin{center}
\includegraphics[scale=1]{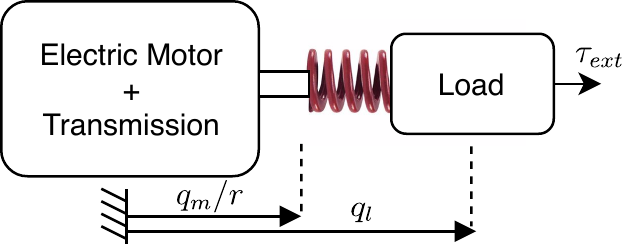}
\end{center}
\caption{Diagram of an SEA. Equations (\ref{eq:EOM_SEA_MotorSide})-(\ref{eq:EOM_SEA_LoadSide}) illustrate the system's equations of motion.}
\label{fig:Esquematic_SEA}
\end{figure}
Figure~\ref{fig:Esquematic_SEA} illustrates the configuration of an SEA. Using the Newton-Euler method, the corresponding balance of torques at the motor and load side provides the following equations of motion
\begin{align}
I_m\ddot{q}_m &= -b_m\dot{q}_m+\tau_m+\frac{\tau_{ela}}{\eta r}, \label{eq:EOM_SEA_MotorSide} \\ 
\tau_{ela} &= g(q_l,\dot{q}_l,\ddot{q}_l,\tau_{ext}), \label{eq:EOM_SEA_LoadSide}
\end{align}
where $I_m$ is the inertia of the motor, $b_m$ its viscous friction coefficient, $\tau_{ela}$ the torque produced by the elastic element, $r$ the transmission ratio, $\eta$ the efficiency of the transmission, and $g(q_l,\dot{q}_l,\ddot{q}_l,\tau_{ext})$ defines the load dynamics as a function of the corresponding load position $q_l$, load velocity $\dot{q}_l$, load acceleration $\ddot{q}_l$, and the external torque applied to the load $\tau_{ext}$ \cite{Spong1987,Verstraten2015, Verstraten2017}. For instance, in the case of an inertial load with viscous friction and an external torque, the load dynamics are defined by $g(q_l,\dot{q}_l,\ddot{q}_l,\tau_{ext}) = -I_l\ddot{q}_l-b_l\dot{q}_l+\tau_{ext}$, where $I_l$ is the inertia of the load, and $b_l$ its corresponding viscous friction coefficient. The variables $q_l,\dot{q}_l,\ddot{q}_l,\tau_{ext}$ represent the given load trajectory. The elongation of the elastic elastic element is defined as $\delta = q_l- q_m/r$. As seen in (\ref{eq:EOM_SEA_MotorSide})-(\ref{eq:EOM_SEA_LoadSide}), the elastic element cannot modify the torque required to perform the motion, $\tau_{ela}$, but it can modify the position of the motor such that $I_m\ddot{q}_m+b_m\dot{q}_m$ reduces the torque of the motor, $\tau_m$.

If periodic motion is considered, $\tau_m$ from (\ref{eq:EOM_SEA_MotorSide}) can be replaced in the expressions of energy (\ref{eq:EnergyConsumedByMotor_IntegralExpression}) to make the following simplification
\begin{align}
&\int_{t_0}^{t_f} \tau_m\dot{q}_m dt = \int_{t_0}^{t_f}\left( I_m\ddot{q}_m+b_m\dot{q}_m-\frac{\tau_{ela}}{\eta r} \right) \dot{q}_m  dt, \nonumber \\
&= \int_{t_0}^{t_f} \left( b_m\dot{q}_m^2-\frac{\tau_{ela}\dot{q}_m}{\eta r} \right) dt + \cancelto{0}{\int_{\dot{q}_{m_0}}^{\dot{q}_{m_f}} I_m \dot{q}_m d\dot{q}_m}, \nonumber \\
&= \int_{t_0}^{t_f} \left( b_m\dot{q}_m^2-\frac{\tau_{ela}}{\eta r}( \dot{q}_m-r\dot{q}_l+r\dot{q}_l) \right) dt, \nonumber \\
&= \int_{t_0}^{t_f} \left( b_m\dot{q}_m^2- \frac{\tau_{ela}\dot{q}_l}{\eta} \right) dt+ \cancelto{0}{\int_{\delta_0}^{\delta_f} \frac{\tau_{ela}}{\eta}  d\delta}, \nonumber \\
&= \int_{t_0}^{t_f} \left( b_m\dot{q}_m^2- \frac{\tau_{ela}\dot{q}_l}{\eta} \right) dt,
\label{eq:SEA_doesNotReduceLoadEnergy}
\end{align}
where $\dot{q}_{m_f} = \dot{q}_{m_0}$ and $\delta_f = \delta_0$ due to periodic motion. This simplification illustrates two concepts. First, canceling the kinetic energy of the motor's rotor for periodic motion will be key to show convexity of energy consumption; when considering non-periodic motion, the expression of energy consumption will not be convex in our optimization framework. Second, the mechanical energy provided to or absorbed from the motion of the load (i.e., $\int_{t_0}^{t_f}\frac{\tau_{ela}\dot{q}_l}{\eta} dt$) is provided or absorbed by the electric motor regardless of the elastic element, i.e., \textit{series elasticity can only reduce the dissipated energy of the motor}. This is relevant for tasks that are mainly dissipative such as level-ground locomotion \cite{Seok2015}.

\subsection{Dynamics of the System in Discrete Form}
Equations (\ref{eq:EOM_SEA_MotorSide})-(\ref{eq:EOM_SEA_LoadSide}) illustrate the dynamics of the system for continuous time derivatives of $q_m \in \mathbb{R}$. In order to formulate this in a convex optimization framework, we approximate the continuous time derivative with a discrete time representation. We discretize time for one period of the reference motion into $n$ points and approximate time derivatives using the following matrix operation $\dot{\bm{q}}_{m}~\approx~D \bm{q}_{m}$, where $\bm{q}_{m}, \dot{\bm{q}}_{m} \in \mathbb{R}^n$ is the discrete-time representation of $q_m$ and $\dot{q}_m$, $D \in \mathbb{R}^{n \times n}$ is
\begin{align}
D &= 
\begin{bmatrix}
			0 & 1 & 0 & 0& \cdots & -1   \\			
			-1 & 0 & 1 & 0 & \cdots & 0 \\
			\vdots & \ddots & & & &	\vdots \\			
			1 & \cdots & & & -1 & 0 \\
\end{bmatrix}\dfrac{1}{2\Delta t},
\end{align}
and $\Delta t$ is the sample rate. $D\bm{q}_{m}$ is the discrete time derivative of $\bm{q}_{m}$ based on the central difference method. The first and last rows of $D$ assume that $\bm{q}_{m}$ represents a periodic trajectory, i.e., $\bm{q}_{m_{(n+1)}}=\bm{q}_{m_{(1)}}$, where $\bm{q}_{m_{(i)}}$ is the $i$th element of the vector $\bm{q}_{m}$. Then the equations of motion, (\ref{eq:EOM_SEA_MotorSide})-(\ref{eq:EOM_SEA_LoadSide}), can be approximated as
\begin{align}
\bm{\tau}_{m} &= (I_mD_2+b_mD)\bm{q}_{m}-\bm{\tau}_{ela}\frac{1}{\eta r}, \label{eq:EOM_discrete_MotorSide}\\
\bm{\tau}_{ela} &= \bm{g}(\bm{q}_{l},\dot{\bm{q}}_{l},\ddot{\bm{q}}_{l},\bm{\tau}_{ext}), \label{eq:EOM_discrete_tauElastic}\\
\bm{\delta} &= \bm{q}_{l}-\bm{q}_{m}\frac{1}{r} \label{eq:EOM_discrete_Elongation},
\end{align}
where $D_2 ~\in~\mathbb{R}^{n\times n}$ is the matrix that computes the second order derivative, and $\bm{\tau}_{m}, \bm{\tau}_{ela}, \bm{q}_{l},\dot{\bm{q}}_{l},\ddot{\bm{q}}_{l},\bm{\tau}_{ext}, \bm{\delta} \in \mathbb{R}^n$ represent the discrete-time versions of the motor torque, torque of the elastic element, load position, load velocity, load acceleration, external torque applied to the load, and elongation of the spring, respectively. The function $\bm{g}(\cdot)$ defines the required values of torque from the elastic element given a reference trajectory, but it does not relate them with the elongation of the spring.

The objective in our optimization is to define the elastic element. This is equivalent to finding the function $\bm{f}(\bm{\delta})$ that relates the given $\bm{\tau}_{ela}$ and the elongation $\bm{\delta}$. Since the elongation is partially defined by the reference load position (\ref{eq:EOM_discrete_Elongation}), the optimization problem can be interpreted as finding the position of the motor, $\bm{q}_{m}$, such that the energy consumption and/or peak power of the motor are minimized. Once the position of the motor is established, the deflection of the elastic element is defined and can be used in conjunction with the given $\bm{\tau}_{ela}$ to generate $\bm{f}(\bm{\delta})$.
\section{Elastic Element Design}\label{sec:ElasticElementDesign}
In this section, we introduce three optimization problems that guide the design of the elastic element. Each optimization problem has a different cost function: energy consumption, peak power, or a combination of these two. However, the three optimization problems share the same constraints to guarantee that a conservative elastic element is obtained and the electric motor is capable to generate the motion. The outcome of each optimization problem is the function $f(\delta)$ in (\ref{eq:EOM_ElasticElementFunction}) to define the elastic element of the SEA. The general idea of the optimization strategy can be summarized as follows:
\begin{align*}
	& \text{minimize}
	&  & \text{\textbf{Energy consumed by the motor}} \\
& & & \text{\textbf{and/or peak power of the motor,}} \\
	& \text{subject to}
   	& & \text{\textit{Actuator constraints.}}
\end{align*}
In this work, convexity of the optimization problem is an important property for the formulation and is exploited whenever possible. In addition to polynomial-time solutions \cite{Nesterov94}, convexity guarantees the existence of a global minimum, and consistent results can be found regardless of the initial point of the optimization solver \cite{Boyd2004}. Regarding the reference motion, the given task is defined as a reference trajectory $q_{l\text{ref}}(t)$, its corresponding time derivatives $\dot{q}_{l\text{ref}}(t)$, $\ddot{q}_{l\text{ref}}(t)$, and the external torque $\tau_{ext}(t)$. In order to draw conclusions independent of the controller design, we assume perfect tracking, i.e., $q_{l}(t) \equiv q_{l\text{ref}}(t)$, $ \dot{q}_{l}(t) \equiv\dot{q}_{l\text{ref}}(t)$, and $\ddot{q}_{l}(t) \equiv \ddot{q}_{l\text{ref}}(t)$.
\subsection{Actuator Constraints}\label{subsec:ActuatorConstraints}
To represent a conservative elastic element, it is sufficient to constrain $f(\delta)$ to be a strictly-increasing monotonic function of $\delta$, i.e., $d\tau_{ela}/d\delta > 0$. In discrete form, $d\tau_{ela}/d\delta~\approx~\Delta\bm{\tau}_{ela}/\Delta\bm{\delta}$, where $\Delta\bm{\tau}_{ela_{(i)}}~=~\bm{\tau}_{ela_{(i)}}-\bm{\tau}_{ela_{(i-1)}}$. Using the definition of elongation (\ref{eq:EOM_discrete_Elongation}), we impose monotonicity using the following equality and inequality constraints:
\begin{align}
A_{1} \bm{q}_m &< \bm{b}_{1}, & A_{2} \bm{q}_m &= \bm{b}_{2}, \label{eq:contraintsA2a}
\end{align}
where the non-zero entries of the sparse matrices $A_1$ and $A_{2}$ are defined as follows:
\begin{align}
&\text{if}\quad\Delta\bm{\tau}_{ela_{(i)}}>0, &
\left\{
\begin{array}{ll}
	A_{1_{(i,i)}} =  1/r\\
	A_{1_{(i,i-1)}} = -1/r\\
	\bm{b}_{1_{(i)}} = \Delta\bm{q}_{l_{(i)}}
\end{array}
  \right.; \nonumber \\
&\text{if}\quad\Delta\bm{\tau}_{ela_{(i)}}<0, &\left\{
\begin{array}{ll}
	A_{1_{(i,i)}} =  -1/r\\
	A_{1_{(i,i-1)}} = 1/r\\
	\bm{b}_{1_{(i)}} = -\Delta\bm{q}_{l_{(i)}}
\end{array}
\right.; \nonumber \\
&\text{if}\quad\Delta\bm{\tau}_{ela_{(i)}}=0, &
\left\{
\begin{array}{ll}
	A_{2_{(i,i)}} =  1/r \\
	A_{2_{(i,i-1)}} =  -1/r \\
	\bm{b}_{2_{(i)}} = \Delta\bm{q}_{l_{(i)}}
\end{array}
\right.; \nonumber
\end{align}
for $i = 2,..., n$. The case of $i = 1$ is defined similarly replacing $i-1$ by $n$, which exploits periodicity of the trajectory. In this work, vector inequalities such as $A_{1} \bm{q}_m < \bm{b}_{1}$ denote componentwise inequalities between vectors, i.e., a generalized inequality with respect to the nonnegative orthant \cite{Boyd2004}.

Our last set of constraints consider the limitations of the electric motor and elastic element. In our framework, maximum torque, maximum angular speed of the electric motor, and maximum elongation of the elastic element can be constrained within the optimization.  Using (\ref{eq:EOM_discrete_MotorSide})-(\ref{eq:EOM_discrete_Elongation}), we impose these limitations as the following vector inequality constraint:
\begin{align}
h(\bm{q}_m) &\leq \bm{u}_{max},
\label{eq:actuatorConstraints}
\end{align}
where 
\begin{align}
h(\bm{q}_m) &= 
\begin{bmatrix}
			\text{max}\{(I_mD_2+b_mD)\bm{q}_{m}-\bm{\tau}_{ela}(\eta r)^{-1}\} \\
			\text{max}\{D\bm{q}_m\} \\
			\text{max}\{\bm{\delta}\}
\end{bmatrix}, \nonumber \\
\bm{u}_{max} &=
\begin{bmatrix}
			\tau_{\text{max}},
			\dot{q}_{\text{max}},
			\delta_{max}
\end{bmatrix}^T, \nonumber 
\end{align}
$\tau_{\text{max}}$, $\dot{q}_{\text{max}}$ are the maximum torque and velocity of the electric motor, and $\delta_{max}$ the maximum elongation of the elastic element. Notice that each element of $h(\bm{q}_m)$ is the composition of the $\text{max}$ function and affine functions of $\bm{q}_m$ \cite[p. 67]{Boyd2004}, which represents a convex function of $\bm{q}_m$ \cite[p. 79]{Boyd2004}. By adding a slack variable, one can rewrite this convex constraint into an affine inequality constraint \cite[pp.~150, 151]{Boyd2004}. This can be used to represent all the constraints in this section as affine equalities and inequalities with respect to $\bm{q}_m$.
\subsection{Minimizing Energy Consumption}
In this section, we use the constraints from Section \ref{subsec:ActuatorConstraints} along with (\ref{eq:EnergyConsumedByMotor_IntegralExpression}) to formulate the convex optimization problem that defines the elastic element that globally minimizes energy consumption of the motor.
\subsubsection{Cost function: Energy consumed by the motor} \label{subsec:EnergyConsumption}
Using the discrete-time formulation of the dynamics (\ref{eq:EOM_discrete_MotorSide})-(\ref{eq:EOM_discrete_Elongation}) and the simplification of rotor mechanical power in (\ref{eq:SEA_doesNotReduceLoadEnergy}), the energy required by the motor (\ref{eq:EnergyConsumedByMotor_IntegralExpression}) can be approximated in discrete form as
\begin{align}
E_m &\approx \sum_{i=1}^{n} \left( \frac{\bm{\tau}_{m_{(i)}}^{2}}{k_m^2}+\bm{\tau}_{m_{(i)}}\dot{\bm{q}}_{m_{(i)}} \right) \Delta t, \nonumber \\
 &= \left( \frac{\bm{\tau}_{m}^T \bm{\tau}_{m}}{k_m^2}+b_m \bm{q}_m^TD^TD\bm{q}_m - \frac{\bm{\tau}_{ela}^T\dot{\bm{q}}_l}{\eta} \right) \Delta t.\label{eq:cost_vectorform}
\end{align}
By defining $\bm{\tau}_m/k_m=F\bm{q}_m+\bm{c}$, where $F=(I_mD_2+b_mD)/k_m$ and $\bm{c}=-\bm{\tau}_{ela}/(\eta k_m r),$ we rewrite (\ref{eq:cost_vectorform}) as
\begin{align}
E_m &= \left(\norm{F\bm{q}_m+\bm{c}}^2_2 + b_m \bm{q}_m^TD^TD\bm{q}_m - {\bm{\tau}_{ela}^T\dot{\bm{q}}_l}/{\eta}\right) \Delta t, \nonumber \\
&= \bm{q}_m^TQ_e\bm{q}_m + A_e\bm{q}_m + c_e,
\label{eq:cost_EnergyQuadratic}
\end{align}
where
\begin{align}
Q_e &= \left( F^TF+b_mD^TD \right) \Delta t, & \label{eq:QuadraticMatrix}\\
A_e &= \left( 2\bm{c}^TF \right) \Delta t,  \\
c_e &= \left( \bm{c}^T\bm{c} - {\bm{\tau}_{ela}^T\dot{\bm{q}}_l}/{\eta} \right) \Delta t.
\end{align}
\subsubsection{Optimization problem} \label{subsec:ConvexProblemEnergyConsumption}
Using the definition of energy in (\ref{eq:cost_EnergyQuadratic}) and the constraints in Section \ref{subsec:ActuatorConstraints}, we write the optimization problem as follows:
\begin{equation}
\begin{aligned}
& \underset{\boldsymbol{q}_{m}}{\text{minimize}}
& & \bm{q}_m^TQ_e\bm{q}_m + A_e\bm{q}_m + c_e,\\
& \text{subject to} & & A_1\boldsymbol{q}_{m} < \boldsymbol{b}_{1},  \\
& & & A_2\boldsymbol{q}_{m} = \boldsymbol{b}_{2},  \\
& & & h(\bm{q}_m) \leq \bm{u}_{max}.  \\
\end{aligned}
\end{equation}
which corresponds to a convex quadratic program (QP) \cite{Boyd2004}. A similar optimization problem was formulated by the authors in the preliminary work \cite{Bolivar2017}.
\subsubsection{Convexity of the optimization problem}
The Hessian of the cost function (\ref{eq:cost_EnergyQuadratic}), $2Q_e$, is a positive semi-definite matrix, therefore the function is convex with respect to $\bm{q}_m$ \cite{Boyd2004}. Positive semi-definiteness can be shown from the definition of $Q_e$ in (\ref{eq:QuadraticMatrix}). The Gramian matrices of $F$ and $D$, i.e., $F^TF$ and $D^TD$, are positive semi-definite as can be seen from their singular value decomposition. The sum of these two matrices is also positive semi-definite. All the constraints can be written as affine functions of $\bm{q}_m$, therefore the optimization problem is convex.
\subsection{Minimizing Peak Power}\label{sec:DesignToMinimizePeakPower}
Design of the elastic element to minimize peak power follows the same principles as in the case of energy consumption. The same constraints apply to both cases, but with a different cost function. The analysis starts with the definition of mechanical power of the electric motor, $\bm{p}_{m}~\in~\mathbb{R}^n$, as a function of $\bm{q}_m$:
\begin{align}
\bm{p}_{m(i)} &= \bm{\tau}_{m(i)}\bm{\dot{q}}_{m(i)}, \nonumber \\
  &= \left( (I_mD_{2(i,*)}+b_mD_{(i,*)})\boldsymbol{q}_{m} -\boldsymbol{\tau}_{ela(i)}\frac{1}{\eta r} \right)D_{(i,*)} \boldsymbol{q}_{m}, \nonumber \\
  &= \boldsymbol{q}_{m}^T \underbrace{\left( I_mD_{2(i,*)}^TD_{(i,*)}+b_mD_{(i,*)}^TD_{(i,*)} \right)}_{G_i} \boldsymbol{q}_{m} \nonumber \\
  & \qquad - \underbrace{\bm{\tau}_{ela(i)}\frac{1}{\eta r} D_{(i,*)}}_{-H_i} \boldsymbol{q}_{m}, \label{eq:DefinitionGiHi} \\
 &= \boldsymbol{q}_{m}^T G_i \boldsymbol{q}_{m} + H_i\boldsymbol{q}_{m},\label{eq:MechanicalPower}
\end{align}
for $i=1, 2,...,n$, where $D_{(i,*)}$ refers to the ith row of the matrix $D \in \mathbb{R}^{n \times n}$. In other words, every element of the power vector, $\bm{p}_m$, is a quadratic expression of the motor position. 

In contrast to energy consumption, the following subsection will show that peak power, $\norm{\boldsymbol{p}_m}_{\infty}$, is not a convex function of $\bm{q}_m$. To keep the advantages of a convex optimization problem, we propose a convex approximation to the expression of peak power. This approximation neglects the torque due to inertia, and maximum power is considered instead of peak power, i.e., we do not take the absolute value. Other approximations of peak power are reported in the literature for numerical optimization, for instance, pseudo-power has been defined in \cite{Hong2017}. Our simulation results indicate that our convex approximation resembles the actual expression of peak power, and that minimizing the convex version minimizes the actual expression as well.
\subsubsection{Optimization problem}\label{subsec:PeakPowerMinimization}
Using the definitions from the previous section, we write the optimization problem as
\begin{equation}
\begin{aligned}
\underset{\boldsymbol{q}_{m}}{\text{minimize}}
& & \norm{\boldsymbol{p}_m}_{\infty},  \\
\text{subject to} & & A_1\boldsymbol{q}_{m} < \boldsymbol{b}_{1},  \\
& & A_2\boldsymbol{q}_{m} = \boldsymbol{b}_{2},  \\
& & h(\bm{q}_m) \leq \bm{u}_{max}.
\end{aligned}
\label{eq:OriginalOptimizationProblemPeakPower}
\end{equation}
Peak power, $\norm{\bm{p}_m}_{\infty}$, is not a convex function of $\bm{q}_m$ because, in general, the infinity norm of a set of quadratic functions is not convex. In addition, every quadratic function may not be convex as shown by its Hessian, $G_i$, in (\ref{eq:DefinitionGiHi}). This matrix is non-definite because the matrix $I_mD_{2(i,*)}^TD_{(i,*)}$ may have positive and negative eigenvalues.

\subsubsection{Simplification to obtain a convex optimization problem}
Peak power can be approximated with maximum power to obtain a convex version of the cost function. The max function, $\text{max}\{f_1, f_2,...,f_n\}$, is convex when each function $f_1, f_2,...,f_n$ is also convex \cite{Boyd2004}. This justifies the use of the max function instead of the infinity-norm. In our case, each function $f_i$ corresponds to the quadratic expression (\ref{eq:MechanicalPower}). These expressions are convex if and only if $G_i$ is positive semi-definite. From its definition (\ref{eq:DefinitionGiHi}), the matrix $G_i$ is positive semi-definite if inertial torques are neglected. With this in mind, we define $\bm{p}^{cvx}_{m} \in \mathbb{R}^n$, a convex approximation of actual power $\bm{p}_m$, as follows:
\begin{align}
\bm{p}^{cvx}_{m(i)} &:= \bm{q}_{m}^T \underbrace{\left( b_mD_{(i,*)}^TD_{(i,*)} \right)}_{G_i^{cvx}} \bm{q}_{m} - \bm{\tau}_{ela(i)}\frac{1}{\eta r} D_{(i,*)} \bm{q}_{m}, \nonumber \\
 &= \bm{q}_{m}^T G_i^{cvx} \bm{q}_{m} + H_i\bm{q}_{m}. \nonumber
\end{align}
With this approximation the convex optimization problem is written as
\begin{equation}
\begin{aligned}
& \underset{\bm{q}_{m}}{\text{minimize}}
& & \text{max}\{\bm{p}^{cvx}_m\},  \\
& \text{subject to} & & A_1\boldsymbol{q}_{m} < \boldsymbol{b}_{1},  \\
& & & A_2\boldsymbol{q}_{m} = \boldsymbol{b}_{2},  \\
& & & h(\bm{q}_m) \leq \bm{u}_{max}.  \\
\end{aligned}
\label{eq:NoInertialOptimizationProblem}
\end{equation}
Convexity can be shown since every element in the vector $\bm{p}^{cvx}_m$ is a convex-quadratic function of $\bm{q}_{m}$, i.e., the matrix $G_i^{cvx}$ is positive semi-definite for all $i$ since it is the Gramian matrix of $D_{(i,*)}$. The maximum of a set of convex functions is also convex \cite[p. 80]{Boyd2004}. All the constraints are affine with respect to $\bm{q}_m$, therefore the optimization problem is convex.

\subsubsection{Regularization to avoid high accelerations} \label{subsec:PeakPowerConvexProblem}
The solution to the optimization problem (\ref{eq:NoInertialOptimizationProblem}) may result in a position trajectory, $\boldsymbol{q}_{m}$, with very high accelerations. This has a great impact in the calculation of the actual peak power, especially for SEAs using a high reduction-ratio transmission. One way to solve this is to penalize solutions with high accelerations. This can be achieved by including acceleration of the motor in the cost function as follows:
\begin{equation}
\begin{aligned}
& \underset{\bm{q}_{m}}{\text{minimize}}
& & \text{max}\{\bm{p}_m^{cvx}\} + \gamma_1 \norm{D_{2}\bm{q}_m}_{\infty},  \\
& \text{subject to} & & A_1\boldsymbol{q}_{m} < \boldsymbol{b}_{1},  \\
& & & A_2\boldsymbol{q}_{m} = \boldsymbol{b}_{2},  \\
& & & h(\bm{q}_m) \leq \bm{u}_{max},  \\
\end{aligned}
\label{eq:RegularizedOptimizationProblem}
\end{equation}
where $\gamma_1$ is a scalar constant that controls the influence of the peak acceleration relative to the maximum of the convex expression of power. Section \ref{sec:Simulation results} elaborates on the selection of $\gamma_1$. The optimization problem (\ref{eq:RegularizedOptimizationProblem}) remains convex. The term $\gamma_1 \norm{D_{2}\bm{q}_m}_{\infty}$ corresponds to the infinity-norm function, which is convex, composed by the linear expression $D_{2}\bm{q}_m$. The composition of a convex function with an affine function results in a convex function \cite{Boyd2004}.

\subsection{Multiobjective Optimization: Energy Consumption and Peak Power}
This section describes the optimization problem that simultaneously minimizes energy consumption and peak power. These two objectives are not always competing; however, when they do, a trade-off curve is useful to guide the design process. For example, a global minimum that reduces energy consumption can lead to higher peak powers compared to actuators without elastic elements. This motivates a multiobjective optimization framework where the designer can choose the appropriate trade-off based on the design specifications. The proposed methodology combines the optimization problems introduced in Sections \ref{subsec:ConvexProblemEnergyConsumption} and \ref{subsec:PeakPowerConvexProblem} to generate the following multiobjective program:
\begin{equation}
\begin{aligned}
& \underset{\bm{q}_{m}}{\text{minimize}}
& & \theta \gamma_2 \left(\bm{q}_m^TQ_e\bm{q}_m + A_e\bm{q}_m + c_e\right) \\
& & & + (1 - \theta)\left(\text{max}\{\bm{p}_m^{cvx}\} + \gamma_1 \norm{D_{2}\bm{q}_m}_{\infty}\right),  \\
& \text{subject to} & & A_1\boldsymbol{q}_{m} < \boldsymbol{b}_{1},  \\
& & & A_2\boldsymbol{q}_{m} = \boldsymbol{b}_{2},  \\
& & & h(\bm{q}_m) \leq \bm{u}_{max}, 
\end{aligned}
\label{eq:ConvexMultiObjectiveOptimization}
\end{equation}
where $\gamma_2$ is a the factor that scales the magnitude of energy consumption relative to peak power, and $\theta \in [0,1]$ is the factor that controls the trade-off between energy consumption and peak power in the solution of the optimization problem. Using $\theta = 0$ indicates that only peak power will be minimized, and $\theta = 1$ minimizes only energy consumption. It can be shown that the solution to the optimization problem in (\ref{eq:ConvexMultiObjectiveOptimization}) corresponds to a pareto optimal point \cite[p.~178]{Boyd2004}. The optimization problem remains convex since it is the positive sum of convex functions.

\section{Case Studies}\label{sec:Simulation results}
In this section, we apply the convex optimization framework in (\ref{eq:ConvexMultiObjectiveOptimization}) to design the SEA's elastic element for two different reference trajectories: natural oscillation of a nonlinear spring, and motion of the human ankle during level-ground walking and running. The first case is of interest because it compares the numerical results of the optimization with the well established analysis of a mass-spring system. This comparison validates the optimization results. For example, if the reference position of the SEA corresponds to the natural oscillation of a nonlinear spring and the cost function is the energy dissipated by viscous friction, the optimal elastic element should be the same as the nonlinear spring used to generate the trajectories. Section \ref{subsec:ExampleCaseNonlinearSpring} illustrates this case and compares it with the optimization results using two different cost functions: winding losses and total energy consumption.

Section \ref{subsec:ExampleCaseAnkle} discusses the design of the SEA's elastic element for the case of the ankle joint of a powered prosthetic leg. For this application, minimizing peak power and energy consumption using SEAs is challenging when considering multiple locomotion tasks. For example, design for level ground walking and running normally leads to different elastic elements for these tasks \cite{Eslamy2012}. In our case study, we design a single nonlinear elastic element that minimizes peak power and energy consumption for both tasks subject to the parameters and constraints for the specific motor-transmission-load system. This illustrates that SEAs with nonlinear elastic elements can be designed for multiple tasks, which is a scenario that traditionally utilizes variable stiffness actuators (VSA) \cite{Wolf2016}. Section \ref{subsubsec:ExtendingRangeofOperation} includes elongation constraints for the elastic element to demonstrate that SEAs using nonlinear springs can perform tasks that a rigid actuator or an SEA with linear springs cannot perform.

For each case study, we used the following configuration. The trajectory of the load, i.e., $q_l, \dot{q}_l, \ddot{q}_l,$ and $\tau_{ext}$ in (\ref{eq:EOM_SEA_LoadSide}), is given and the optimization problem is numerically solved using {CVX}, a package for specifying and solving convex programs \cite{cvx,Grant2008}. In all simulations, CVX executed the solver Mosek \cite{Aps2017} with precision settings~\verb|cvx_precision best|. For the two case studies, we used the parameters of a commercial frameless motor (Model: ILM 85x26, RoboDrive, Seefeld, Germany, Table~\ref{table:SimulationParameters.}), motivated by the motor selection of the second-generation of the powered prosthetic leg at the University of Texas at Dallas \cite{Elery2018}. This motor has a high nominal and peak torque, requiring a lower reduction ratio. This configuration favors backdrivability of the SEA.

\begin{table}
\renewcommand{\arraystretch}{1.3}
\caption{Simulation parameters: motor ILM85x26 from RoboDrive.}
\label{table:SimulationParameters.}
\centering
\begin{tabular}{lcc}
\hline
Parameter	&ILM85x26 & Units\\
\hline
Motor torque constant, $k_t$ & 0.24 & \SI[inter-unit-product =$\cdot$]{}{\newton\meter\per\ampere}\\
Motor terminal resistance, $R$ & 323 & \SI{}{\milli\ohm} \\
Rotor inertia, $I_{mr}$ & 1.15 & \SI{}{\kilo\gram\square\centi\meter}\\
Rotor assembly, $I_{ma}$ & 0.131 & \SI{}{\kilo\gram\square\centi\meter}\\
Motor inertia, $I_m=I_{mr}+I_{ma}$ & 1.246 & \SI{}{\kilo\gram\square\centi\meter}\\
Gear ratio, $r$ & 22 & \\
Motor viscous friction, $b_m$ & 60 & \SI{}{\micro\newton\meter\second\per\radian}\\
Max. motor torque, $\tau_{max}$ & 8.3 & \SI{}{\newton\meter}\\
Max. motor velocity, $\dot{q}_{max}$ & 1500 & \SI{}{rpm}\\
Nominal power output, @\SI{48}{\volt} & 410 & \SI{}{\watt}\\
Peak power output, @(\SI{48}{\volt}, $\tau_{max}$) & 1259 & \SI{}{\watt}\\
\hline
\end{tabular}
\end{table}

\subsection{Natural Oscillation of a Nonlinear Spring}\label{subsec:ExampleCaseNonlinearSpring}

This case study analyzes the design of the elastic element to minimize energy consumption, which can be separated into the following cost functions: energy associated with viscous friction, winding heat losses, and the total dissipated energy. In this section, we solve the optimization problem for each of these cost functions. 
\begin{figure}[t]
\begin{center}
\includegraphics[scale=1]{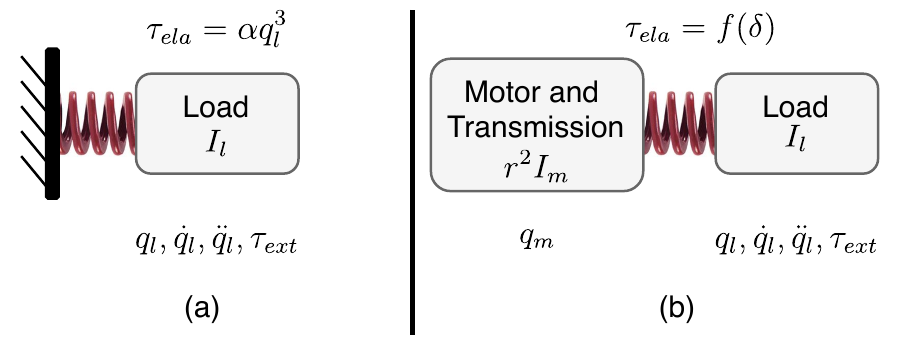}
\end{center}
\caption{(a) Single-mass spring system. The elastic element describes the nonlinear spring with $\tau_e=\alpha q_l^3$. (b) Double-mass single-spring system. The equilibrium position of the elastic element is $q_l=q_m/r$, elongation is defined as $\delta = q_l-q_m/r$. Motor and transmission are considered to be backdrivable.}
\label{fig:Esquematic_ThirdOrderSpring} 
\end{figure}
Figure~\ref{fig:Esquematic_ThirdOrderSpring}-(a) describes a single mass-spring system with a nonlinear spring, $\tau_{ela}=\alpha q_l^3$, and corresponding equation of motion $\tau_{ela}=-I_l\ddot{q}_l$, where $I_l=$ \SI{125}{\gram\square\meter} is the inertia of the load and $\alpha=$ \SI{40}{\newton\meter\per\cubic\radian}. The main objective of this case study is to validate the optimization results against the known natural-oscillation of a cubic spring. To simplify the analysis, we consider no actuator constraints and a no-loss mechanical transmission, i.e., $\eta=1$. Section \ref{subsec:ExampleCaseAnkle} describes the design with actuator constraints and inefficiencies in the transmission. Figure~\ref{fig:MotionLoad3order} illustrates the motion of the load for an initial displacement, $q_l(0)=\SI{\pi/2}{\radian}$. This natural vibration is defined as our reference motion. The SEA, Fig.~\ref{fig:Esquematic_ThirdOrderSpring}-(b),  can generate this motion with the motor holding its initial position if the elastic element matches the nonlinear spring in Fig.~\ref{fig:Esquematic_ThirdOrderSpring}-(a). However, this approach may not be energetically efficient. If the load is high enough to backdrive the system, the motor must apply a reactionary torque to hold its initial position. This torque requires a current that generates heat losses at the motor's winding due to Joule heating. In contrast, we can solve the optimization problem in (\ref{eq:ConvexMultiObjectiveOptimization}) to find the elastic element that minimizes the total energy expenditure (i.e., winding losses and viscous friction). To evaluate the proposed methodology, we solved the optimization problem for each of the following cost functions: energy dissipated by winding Joule heating, energy dissipated by viscous friction, and total energy consumption. Each of these cost functions is formulated from appropriate modifications to the matrices $Q_e$ and $A_e$ in (\ref{eq:cost_EnergyQuadratic}). The resulting elastic elements, torques, and positions of the motor are illustrated in Fig.~\ref{fig:optimalElasticElementsThirdOrder}. Table~\ref{table:SimulationResultsNaturalOscillation} summarizes the energy expenditure for each case.
\begin{figure}[h!]
\setlength{\abovecaptionskip}{5pt}
\begin{center}
\begin{tikzpicture}
  \begin{axis}[
    legend style = {
    legend pos = outer north east,
    fill = none,
    draw = none,
    },
  	width = 0.3\textwidth,
  	height = 0.11\textheight,
    scale only axis,    
    xlabel= Time \lbrack\SI{}{\milli\second}\rbrack,,
    ylabel= $\tau_{ext}$ \lbrack\SI{}{\newton\meter}\rbrack]
  \addplot [blue!50, solid ,line width=1pt] table[x=Time,y=TorqueLoad]
	{NonlinearResults_CostFn_TotalEnergy.dat} coordinate [pos=0.55] (tauExtCoor);
	\node [
		coordinate,
		pin = right:{$\tau_{ext}$},
	] 
		at (tauExtCoor) {}
	;
	\label{plot_one}
  \end{axis}
  \begin{axis}[
		legend style = {
	    legend pos = outer north east,
	    fill = none,
	    draw = none,
	    },  
	    width = 0.3\textwidth,
	  	height = 0.11\textheight,
	  	scale only axis, 
	    axis y line*=right,
	    axis x line=none,
	    ylabel= $q_l$ \lbrack\SI{}{\radian}\rbrack,
    ]
    \addplot [red!50, dashed, line width=1pt] table[x=Time,y=LoadPosition]{NonlinearResults_CostFn_TotalEnergy.dat} coordinate [pos=0.55] (qlCoor);
	\node [
		coordinate,
		pin = right:{$q_l$},
	] 
		at (qlCoor) {}
	;
	
  \end{axis}
\end{tikzpicture}
\end{center}
\caption{ The reference trajectory of the load is defined by the natural oscillation of the single mass-spring system in Fig.~\ref{fig:Esquematic_ThirdOrderSpring}-(a) with $\alpha= $\SI{40}{\newton\meter\per\cubic\radian}, $I_l= $\SI{125}{\gram\square\meter}, and $q_l(0)=\SI{\pi/2}{\radian}$.}
\label{fig:MotionLoad3order}
\end{figure}
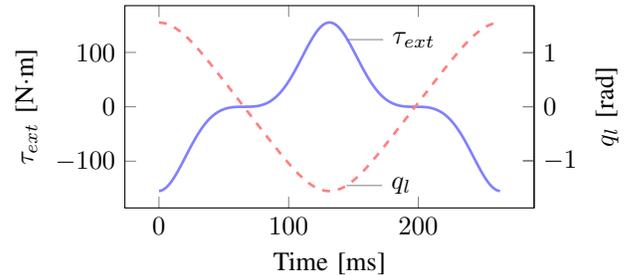
\begin{figure}[h!]
\setlength{\abovecaptionskip}{5pt}
\begin{tikzpicture}
	\begin{groupplot}[
		group style={
			group size=1 by 3,
			vertical sep = 1.3cm,			
			xlabels at = edge bottom,			
			},
		width = 0.45\textwidth, 
		height = 0.19\textheight,		
		xlabel = Time \lbrack\SI{}{\milli\second}\rbrack,		
		]
	\nextgroupplot [
		xlabel = $\delta$ \lbrack\SI{}{\radian}\rbrack,
		ylabel  = $\tau_{ela}$ \lbrack\SI{}{\newton\meter}\rbrack,
		legend style = {
		font=\footnotesize,
    	legend pos = south east,
    	fill = none,
    	draw = none,
    	},
		legend entries = {Viscous friction, Winding losses, Total energy},
		]
	\addplot [red!50, solid, line width=1pt ] table[x=Elongation,y=TorqueElastic]{ViscousFriction_TorqueElastic.dat} coordinate [pos=0.90] (VFric);
	\node [
		coordinate,
		pin = left:{\footnotesize $\approx 40\delta^3$},
	] 
		at (VFric) {}
	;
	\addplot [blue!50, dashed, line width=0.9pt ] table[x=Elongation,y=TorqueElastic]{JouleHeating_TorqueElastic.dat};
	\addplot [gray, loosely dotted, line width=1pt] table[x=Elongation,y=TorqueElastic]{TotalEnergyConsumption_TorqueElastic.dat};
	\nextgroupplot [
		xlabel = Time \lbrack\SI{}{\milli\second}\rbrack,
		ylabel  = $\tau_m$ \lbrack\SI{}{\newton\meter}\rbrack,
		legend pos = south west,
		height = 0.19\textheight,
		]
	\addplot [red!50, solid, line width=1pt ] table[x=Time,y=TorqueMotor]{ViscousFriction_TorqueMotor.dat};
	\addplot [blue!50, dashed, line width=1pt ] table[x=Time,y=TorqueMotor]{JouleHeating_TorqueMotor.dat};
	\addplot [gray, loosely dotted, line width=1pt ] table[x=Time,y=TorqueMotor]{TotalEnergyConsumption_TorqueMotor.dat};		
	\nextgroupplot [
		xlabel = Time \lbrack\SI{}{\milli\second}\rbrack,
		ylabel  = $q_m$ \lbrack\SI{}{\radian}\rbrack,
		legend pos = south west,
		height = 0.19\textheight,
		]
	\addplot [red!50, solid, line width=1pt ] table[x=Time,y=PositionM]{ViscousFriction_PositionM.dat};
	\addplot [blue!50, dashed, line width=1pt ] table[x=Time,y=PositionM]{JouleHeating_PositionM.dat};
	\addplot [gray, loosely dotted, line width=1pt ] table[x=Time,y=PositionM]{TotalEnergyConsumption_PositionM.dat};	
	\end{groupplot}
\end{tikzpicture}
\caption{Optimization results considering natural oscillation of a nonlinear spring as the reference motion. The solid line corresponds to the elastic element that minimizes the energy consumption due to viscous friction. It matches $\tau_{ela}=40\delta^3$, the nonlinear spring used in the single mass-spring system. The dotted line describes the elastic element that minimizes winding losses due to Joule heating. The dashed line describes the elastic element that minimizes both winding losses and viscous friction, i.e., total energy. The corresponding energy expenditure is shown in Table~\ref{table:SimulationResultsNaturalOscillation}.}
\label{fig:optimalElasticElementsThirdOrder}
\end{figure}
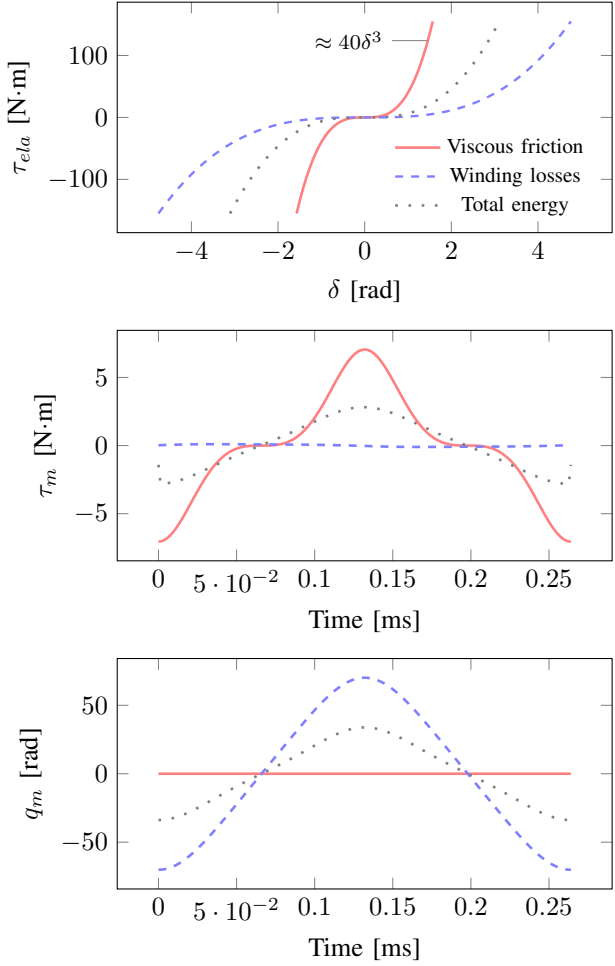
\begin{table}[t]
\renewcommand{\arraystretch}{1.3}
\caption{Energy expenditure. Natural oscillation of cubic spring.}
\label{table:SimulationResultsNaturalOscillation}
\centering
\begin{tabular}{cccc}
\hline
Cost 	& Joule   &  Viscous  & Total  \\
 function & heating \lbrack\SI{}{\joule}\rbrack & friction \lbrack\SI{}{\joule}\rbrack & Energy \lbrack\SI{}{\joule}\rbrack \\
\hline
Viscous friction & 20.175  & 0.000 & 20.175  \\
Winding losses & 0.008  &  20.511  & 20.519  \\
Total energy & 4.972  & 4.607  & 9.579 \\
\hline
\end{tabular}
\end{table}

Minimizing viscous friction leads to the same elastic element as in Fig.~\ref{fig:Esquematic_ThirdOrderSpring}-(a), which validates the numerical results with respect to first principles. The energy required to produce the motion is then \SI{20.175}{\joule}, which is all dissipated in the motor's winding. In contrast, minimizing the total energy consumption results in a cost of \SI{9.579}{\joule}, 52\% less compared to the previous case. The elastic element is nonlinear but is not defined by $\tau_{ela}=40\delta^3$, and the motor no longer remains stationary. Minimizing only the energy dissipated by the motor's winding leads to an elastic element that minimizes as much motor-torque as possible, as seen in Fig.~\ref{fig:optimalElasticElementsThirdOrder}. 
This SEA spring approximates the natural dynamics of the double-mass single-spring system defined by the inertia of the load and the motor.

\subsection{Ankle Joint of a Powered Prosthetic Leg}\label{subsec:ExampleCaseAnkle}
This section presents the design of the elastic element of an SEA for the ankle joint of a powered prosthetic leg. In this case study, we constrain the maximum absolute value of torque and velocity of the motor to be within the specifications of the datasheet (Table \ref{table:SimulationParameters.}). In Section \ref{subsubsec:unconstrainedElongation}, we design the elastic element without constraining its elongation. In Section \ref{subsubsec:ExtendingRangeofOperation}, we constrain the maximum elongation of the spring to show that nonlinear springs can extend the range of operation of SEAs, i.e., allow SEAs to perform tasks that are not possible with linear springs or rigid actuators. These sections analyze a system with energy losses at the transmission, i.e., $\eta = 0.8$.
\subsubsection{Unconstrained elongation}\label{subsubsec:unconstrainedElongation}
\begin{figure}
\setlength{\abovecaptionskip}{5pt}
\begin{tikzpicture}
	\begin{groupplot}[
		group style={
			group size=1 by 2,
			xlabels at = edge bottom,	
			vertical sep = 0.4cm,		
			},
		width = 0.45\textwidth, 
		height = 0.19\textheight,		
		xlabel = Percentage of gait cycle \lbrack \%\rbrack,		
		]
	\nextgroupplot [
		ylabel  = $q_l$ \lbrack\SI{}{\radian}\rbrack,
		legend entries = {Walking, Running}, 
		legend pos = north east,
		legend style={
			draw=none,
			font=\footnotesize,
			},
		xtick = {},
		xticklabels = \empty,
		]
	\addplot [blue!50, solid, line width=1pt] table[x = PercentageGC, y = LoadPosition]{refTrajectory_75kg_level_walking.dat};
	\addplot [red!50, dashed, line width=1pt] table[x = PercentageGC, y = LoadPosition]{refTrajectory_75kg_running.dat};
	\nextgroupplot [
		ylabel  = $\tau_{ext}$ \lbrack\SI{}{\newton\meter}\rbrack,
		]
	\addplot [blue!50, solid, line width=1pt,] table[x = PercentageGC, y = LoadTorque]{refTrajectory_75kg_level_walking.dat};	
	\addplot [red!50, dashed, line width=1pt,] table[x = PercentageGC, y = LoadTorque]{refTrajectory_75kg_running.dat};	
	\end{groupplot}
\end{tikzpicture}
\caption{Motion of the human ankle during level ground walking \cite{Winter91} and running \cite{Novacheck1998}. The gait cycle begins with heel contact of one foot and finishes with the subsequent occurrence of the same foot. In the lower figure, the external torque, $\tau_{ext}$, is defined for a \SI{75}{\kilo\gram} subject.}
\label{fig:referenceMotionWalkingRunningIndependent}
\end{figure}
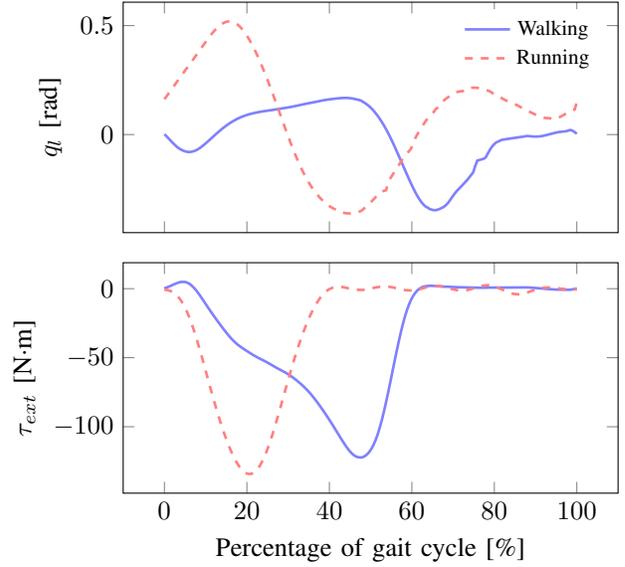

One of the advantages of the proposed methodology is the capability to analyze arbitrary periodic reference trajectories. Taking advantage of this flexibility, the design of the elastic element considers three different tasks for the prosthetic leg: level-ground walking or running as shown in Fig.~\ref{fig:referenceMotionWalkingRunningIndependent}, and a combination of walking and running. The walking and running trajectory combines four steps of walking and one of running, which corresponds to the case where the user runs 20\% of the time and walks during the remaining portion of operation.

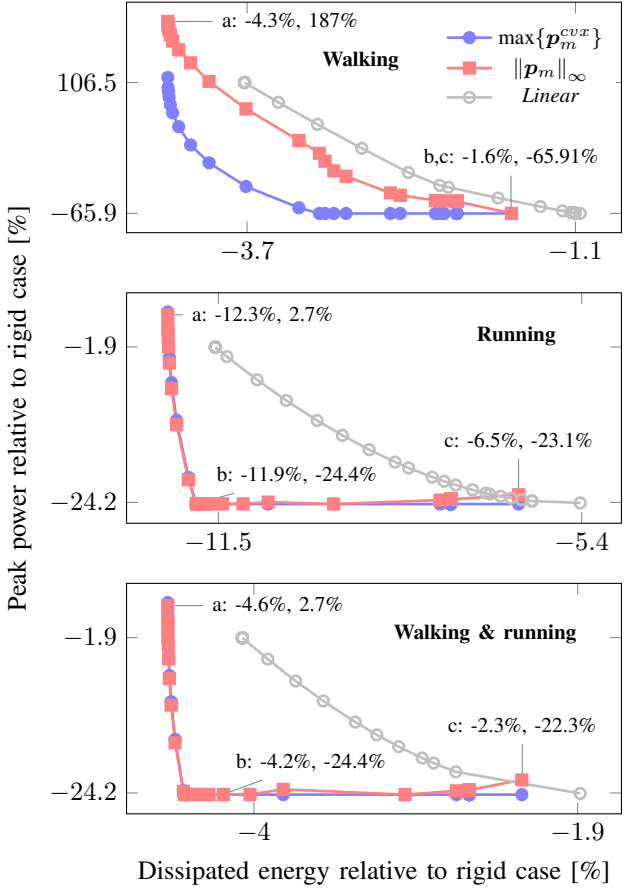
\begin{figure}[h!]
\setlength{\abovecaptionskip}{5pt}
\begin{tikzpicture}
\begin{groupplot}[
	group style = {
			group size = 1 by 3,
			vertical sep = 0.8cm,
			xlabels at = edge bottom,			
			},
	width = 0.45\textwidth,
	height = 0.19\textheight,
	xlabel = {Dissipated energy relative to rigid case [\%]},
	]
	\nextgroupplot[
		legend entries = {$\text{max}\{\bm{p}_m^{cvx}\}$, $\norm{\boldsymbol{p}_m}_{\infty}$, \textit{Linear}}, 
		legend pos = north east,
		legend style={draw=none,at={(.99,.5)},anchor=south east, font=\footnotesize,},	
		ylabel near ticks,
		yticklabel pos=left,
		xtick = {-3.7, -1.1},
		ytick = {106.5, -65.9},
		]
		\addplot [blue!50, mark = *, mark color =blue!50, line width = 1pt] table[x = GeneralEnergyDissipated, y = GeneralCVXPeakPower]{180501_TMECH_Results_Ankle_Walking_TradeOff.dat};
		\addplot [red!50, mark = square*, mark color =red!50, line width = 1pt] table[x = GeneralEnergyDissipated, y = GeneralPeakPower]{180501_TMECH_Results_Ankle_Walking_TradeOff.dat};
		\addplot [gray!50, mark = o, mark color =gray, line width = 1pt] table[x = LinearEnergyDissipated, y = LinearPeakPower]{180501_TMECH_Results_Ankle_Walking_TradeOff.dat};
		\node [coordinate, pin ={[pin distance=0.6cm]0: \footnotesize a: -4.3\%, 187\%}] at (axis cs: -4.33, 187.02){};
		\node [coordinate, pin ={[pin distance=0.5cm]90: \footnotesize b,c: -1.6\%, -65.91\%}] at (axis cs: -1.61, -65.91){};
		\node[above] at (axis cs: -2.8,110) {\footnotesize \textbf{Walking}};
	\nextgroupplot[
		ylabel  = {Peak power relative to rigid case [\%]},
		legend pos = north east,
		legend style={draw=none,at={(.99,.5)},anchor=south east, font=\footnotesize,},
		ylabel near ticks,
		yticklabel pos=left,
		xtick = {-11.5, -5.4},
		ytick = {-1.9, -24.2},
		]
		\addplot [blue!50, mark = *, mark color =blue!50, line width = 1pt] table[x = GeneralEnergyDissipated, y = GeneralCVXPeakPower]{180501_TMECH_Results_Ankle_Running_TradeOff.dat};
		\addplot [red!50, mark = square*, mark color =red!50, line width = 1pt] table[x = GeneralEnergyDissipated, y = GeneralPeakPower]{180501_TMECH_Results_Ankle_Running_TradeOff.dat};
		\addplot [gray!50, mark = o, mark color =gray, line width = 1pt] table[x = LinearEnergyDissipated, y = LinearPeakPower]{180501_TMECH_Results_Ankle_Running_TradeOff.dat};
		\node[above] at (axis cs: -6.5,-3) {\footnotesize \textbf{Running}};
		\node [coordinate, pin={[pin distance=0.2cm]0: \footnotesize a: -12.3\%, 2.7\%}] at (axis cs: -12.35, 2.71){};
		\node [coordinate, pin ={[pin distance=0.2cm]45: \footnotesize b: -11.9\%, -24.4\%} ] at (axis cs: -11.86, -24.41){};
		\node [coordinate, pin=above:{\footnotesize c: -6.5\%, -23.1\%}] at (axis cs: -6.46, -23.10){};
\nextgroupplot[
		legend pos = north east,
		legend style={draw=none,at={(.99,.5)},anchor=south east, font=\footnotesize,},
		ylabel near ticks,
		yticklabel pos=left,
		xtick = {-4, -1.9},
		ytick = {-1.9, -24.2}
		]
		\addplot [blue!50, mark = *, mark color =blue!50, line width = 1pt] table[x = GeneralEnergyDissipated, y = GeneralCVXPeakPower]{180501_TMECH_Results_Ankle_WalkingAndRunning_TradeOff.dat};
		\addplot [red!50, mark = square*, mark color =red!50, line width = 1pt] table[x = GeneralEnergyDissipated, y = GeneralPeakPower]{180501_TMECH_Results_Ankle_WalkingAndRunning_TradeOff.dat};
		\addplot [gray!50, mark = o, mark color =gray, line width = 1pt] table[x = LinearEnergyDissipated, y = LinearPeakPower]{180501_TMECH_Results_Ankle_WalkingAndRunning_TradeOff.dat};
		\node[above] at (axis cs: -2.5,-4) {\footnotesize \textbf{Walking \& running}};
		\node [coordinate, pin=right:{\footnotesize a: -4.6\%, 2.7\%}] at (axis cs: -4.56, 2.69){};
		\node [coordinate, pin ={[pin distance=0.2cm]75:\footnotesize b: -4.2\%, -24.4\%} ] at (axis cs: -4.2, -24.42){};
		\node [coordinate, pin=above:{\footnotesize c: -2.3\%, -22.3\%}] at (axis cs: -2.26, -22.33){};
\end{groupplot}
\end{tikzpicture}
\caption{Trade-off curves for the tasks of level ground walking, running, and walking and running. Point (a) in each graph indicates the results when the cost function is only energy consumption, i.e., $\theta = 1$ in (\ref{eq:ConvexMultiObjectiveOptimization}). Point (b) represents the optimal point based on the trade-off analysis. Point (c) represents the results when the cost function is only maximum convex power, i.e., $\theta = 0$ in (\ref{eq:ConvexMultiObjectiveOptimization}). Relative percentage is computed as $100(x_{optim} - x_{rigid})/x_{rigid}$, where $x_{optim}$ and $x_{rigid}$ are the dissipated energy or peak power from the optimization algorithm and rigid case respectively. Peak power and its corresponding convex approximation are denoted by $\norm{\boldsymbol{p}_m}_{\infty}$ and $\text{max}\{\bm{p}_m^{cvx}\}$ respectively. As a reference, we include the energy and peak power savings using a linear series spring (legend: \textit{Linear}).}
\label{fig:tradeoffCurve}
\end{figure}

The multiobjective optimization involves analysis of the trade-off curves in Fig.~\ref{fig:tradeoffCurve}. Results are reported relative to the peak power and dissipated energy of a rigid actuator performing the same task. The dissipated energy with a rigid actuator is \SI{59.27}{\joule} and \SI{33.75}{\joule} per gait cycle for walking and running tasks, respectively, while its peak power reaches \SI{325.29}{\watt} and \SI{1111.12}{\watt} for these tasks. Note that the dissipated energy values are per cycle, and because the walking period is longer than running (\SI{1.14}{\second} walking \cite{Winter91} and  \SI{0.66}{\second} for running \cite{Novacheck1998}), it dissipates more energy, even though the running peak torques are higher (Fig. \ref{fig:referenceMotionWalkingRunningIndependent}).

To generate the trade-off curves in Fig.~\ref{fig:tradeoffCurve}, {CVX} solved the optimization problem (\ref{eq:ConvexMultiObjectiveOptimization}) using 30 different values for $\theta \in [0,1]$. The points between $\theta = 0$ and $\theta = 1$ were sampled from a sigmoid function to have an adequate distribution of points in the trade-off curves. In the proposed methodology, $\gamma_1$ and $\gamma_2$ in (\ref{eq:ConvexMultiObjectiveOptimization}) control the relative magnitude of the two costs. $\gamma_1$ scales the maximum acceleration with respect to the convex simplification of peak power and $\gamma_2$ scales energy consumption relative to peak power. Comparing the relative magnitude of peak power and energy consumption, we defined $\gamma_1 = 0.02$ and $\gamma_2 = 300$.

\begin{figure}[h!]
\setlength{\abovecaptionskip}{5pt}
\begin{tikzpicture}
	\begin{groupplot}[
		group style={
			group size=1 by 1,
			xlabels at = edge bottom,			
			},
		width = 0.45\textwidth, 
		height = 0.18\textheight,		
		xlabel = {Elongation elastic element, $\delta$, \lbrack\SI{}{\radian}\rbrack},		
		]
	\nextgroupplot [
		axis y line*=left,
	    axis x line*=bottom,
		ylabel  = $\tau_{ela}$ \lbrack\SI{}{\newton\meter}\rbrack,
		]
	\addplot [blue!50, solid, line width=1pt] table[x = elongationGeneral75kg, y = torqueElasticGeneral75kg]{180501_TMECH_Results_Ankle_Walking_ElasticElement.dat};	
	\addplot [red!50, dashed, line width=1pt] table[x = elongationGeneral75kg, y = torqueElasticGeneral75kg]{180501_TMECH_Results_Ankle_Running_ElasticElement.dat};
	\addplot [gray!50, loosely dashdotdotted, line width=1pt] table[x = elongationGeneral75kg, y = torqueElasticGeneral75kg]{180501_TMECH_Results_Ankle_WalkingAndRunning_ElasticElement.dat};
	\node [coordinate, pin=left:{\footnotesize Walking}] at (axis cs: 0.330740, 89.652440){}; 
	\node [coordinate, pin=right:{\footnotesize Running}] at (axis cs: 0.301012, 31.746396){}; 
	\node [coordinate, pin=left:{\footnotesize Walking \& running}] at (axis cs: 0.893581, 134.265709){}; 
	\end{groupplot}
\end{tikzpicture}
\caption{Each elastic element corresponds to the solution described by the optimal points (b) in Fig.~\ref{fig:tradeoffCurve}, i.e., represents the optimal point based on our trade-off analysis for each trajectory. Each solution weights energy consumption and peak power differently as the value of $\theta$ is different in each case.}
\label{fig:OptimalElasticElement}
\end{figure}
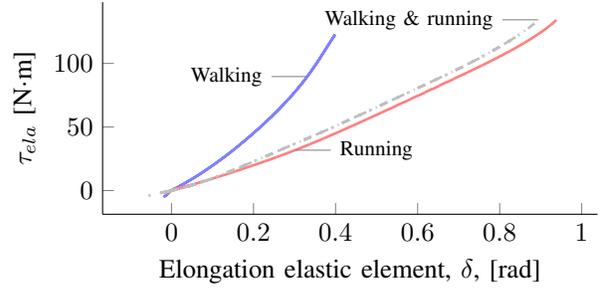

For the given trajectories and motor configuration, there is a correlation between reduction of convex power and peak power, as shown in Fig.~\ref{fig:tradeoffCurve}. In addition, the relationship between energy consumption and peak power represents a \textit{weak trade-off} in the multiobjective optimization \cite[p. 182]{Boyd2004}, i.e., a small increase in the dissipated energy will imply a significant reduction of peak power. For example, in the walking gait trade-off curve of Fig.~\ref{fig:tradeoffCurve}, point (a) represents a reduction of 4.3\% of dissipated energy and an increase of 187\% of peak power when compared to a rigid actuator. A significant reduction of peak power is achieved by consuming a little more of energy, as seen in point (b) where dissipated energy and peak power reduce 1.61\% and 65.91\% respectively. The other two tasks use the same principle to define point (b), the solution of the multiobjective optimization problem. The selection of this point depends on the priorities of the designer. For example, if energy consumption is the main cost to minimize, then point (a) should be selected. For the case of peak power, point (c) may represent the best choice of the elastic element. Point (c) is defined with respect to the convex approximation of peak power, $\text{max}\{\bm{p}_m^{cvx}\}$. Thus, it is possible to find optimal solutions with lower peak power and $\theta \neq 0$, e.g., point (b) for running and walking \& running. Figure~\ref{fig:OptimalElasticElement} illustrates the optimal elastic element for each of the optimal trade-off points (b). These conservative elastic elements are nonlinear and satisfy the constraints defined in Section~\ref{subsec:ActuatorConstraints}. 
\subsubsection{Constrained elongation}\label{subsubsec:ExtendingRangeofOperation}
\begin{figure}[h!]
\setlength{\abovecaptionskip}{5pt}
\begin{tikzpicture}
	\begin{groupplot}[
		group style={
			group size=1 by 1,
			xlabels at = edge bottom,			
			},
		width = 0.45\textwidth, 
		height = 0.19\textheight,		
		xlabel = {Elongation elastic element, $\delta$, \lbrack\SI{}{\radian}\rbrack},		
		]
	\nextgroupplot [
		ylabel  = $\tau_{ela}$ \lbrack\SI{}{\newton\meter}\rbrack,
		]
	\addplot [blue!50, solid, line width=1pt] table[x = elongationGeneral75kg, y = torqueElasticGeneral75kg]{180501_TMECH_Results_Ankle_RunningElongConstr_ElasticElement.dat};
	\node [coordinate, pin=right:{\footnotesize \SI{2562}{\kilo\newton\meter\per\radian}}] at (axis cs: 0.00010117, 7.3892){};
	\node [coordinate, pin=right:{\footnotesize \SI{303}{\newton\meter\per\radian}}] at (axis cs: 0.1796, 95.5657){};
	\end{groupplot}
\end{tikzpicture}
\caption{Optimal nonlinear elastic element subject to elongation and motor constraints. Local values of stiffness are reported in the graph. At small elongations the elastic element is almost rigid (\SI{2562}{\kilo\newton\meter\per\radian}). A parametric representation of this elastic element would involve a polynomial of high degree, which may be cumbersome for existing methods of design.}
\label{fig:OptimalElasticElementConstrained}
\end{figure}
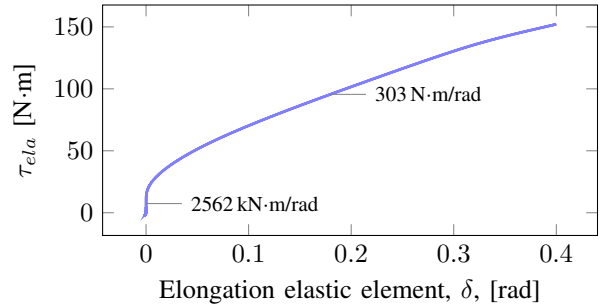

In this section, we design the series elastic element constraining its maximum elongation. This exemplifies the case where nonlinear series elasticity can achieve tasks that linear SEAs or rigid actuators cannot accomplish. In this case study, the task is defined by the kinematics and kinetics for the ankle joint of an \SI{85}{\kilo\gram} subject during running \cite{Novacheck1998}. In addition to the constraints in torque and velocity of the motor (less than \SI{8.3}{\newton\meter} and \SI{1500}{rpm} respectively, Table~\ref{table:SimulationParameters.}), we limit the maximum elongation of the spring to be less than \SI{0.4}{\radian}. This constraint may be imposed by the geometry of the mechanism or maximum elongation of the spring. For the rigid actuator, the task will require a peak torque of \SI{9}{\newton\meter} and maximum absolute speed of \SI{1674}{rpm}, which is outside the motor's specifications using a \SI{48}{\volt} power source. The solution of the linear spring approaches the characteristics of the rigid actuator as we constrain the elongation. In contrast, the nonlinear spring in Fig.~\ref{fig:OptimalElasticElementConstrained} elongates less than \SI{0.4}{\radian}, while the motor torque and velocity remain within specifications. The dissipated energy using the optimal nonlinear spring is \SI{40.67}{\joule} per cycle and the peak power is \SI{1161.96}{\watt}.
\section{Discussion and Conclusion}\label{sec:ConclAndDiscuss}

The proposed framework formulates the design of elastic elements for SEAs as a convex optimization problem to minimize peak power and energy consumption. The formulation considers arbitrary periodic trajectories, actuator constraints, and a non-parametric description of the torque-elongation relationship of the elastic element. The case studies illustrate practical implications of the trade-off analysis between peak power and energy consumption, the accuracy of the convex approximation of peak power, and how to increase the range of operation of SEAs using nonlinear springs.

The trade-off between peak power and energy consumption depends on the characteristics of the load, motor, and transmission. For instance, for the results in Section \ref{sec:Simulation results}, minimizing peak power produced a reduction in both peak power and energy consumption for all the cases studied. However, minimizing energy alone increased peak power. Using a high-speed low-torque motor, as in \cite{Bolivar2017}, had the opposite effect, where minimizing energy consumption also decreased peak power. Though results depend on the characteristics of the load, motor, and transmission, our simulation results show the importance of considering peak power for design, as seen by the weak trade-off between peak power and energy consumption in Fig.~\ref{fig:tradeoffCurve}. Similar results are reported in \cite{Grimmer2011} using linear springs.

Convexity of energy consumption indicates that the optimal point corresponds to a global optimum, which is exploited to obtain the maximum amount of energy that can be reduced using SEAs. This upper bound can also guide the multiobjective version of the optimization. The designer can trade an increase in energy consumption for a reduction of peak power using as a reference the maximum amount of energy that can be reduced. For example, for the given motor-transmission configuration during level ground walking, the maximum amount of energy that can be reduced is about 4.3\% with a corresponding increase of 187\% of peak power. Reducing dissipated energy to 1.61\% correlates to a decrease of 65.91\% of peak power. Saving 1.61\% of the dissipated energy may seem small, but comparing it with the maximum savings of 4.3\% gives a better perspective.

The convex approximation of peak power is close to the actual expression of power depending on the motor configuration. For all the cases considered in this article (Fig.~\ref{fig:tradeoffCurve}), the approximation was accurate enough to provide a significant reduction of the actual expression of peak power. However, using a high transmission ratio increases the relevance of inertial torques in the definition of peak power. In this case the convex approximation is less likely to yield an accurate estimation. The designer can evaluate the performance of the approximation offline to guide the design process. Using the convex approximation has significant advantages with respect to the quality of the solution and the time required to solve the optimization problem. In this case, the proposed convex QCQP program can be efficiently solved in polynomial-time \cite{Nesterov94}, which is useful for an actuator that can modify its stiffness during operation, e.g., a VSA.

The capability to analyze arbitrary periodic trajectories is useful for the design of spring profiles that minimize energy consumption and peak power for a variety of tasks, e.g., the walking and running trajectories in Section \ref{subsubsec:unconstrainedElongation}. In addition, nonlinear SEAs can accomplish tasks that are impossible for linear SEAs or rigid actuators, as shown in Section \ref{subsubsec:ExtendingRangeofOperation}. This property of nonlinear springs and the possibility to analyze any set of periodic trajectories enables our methodology to design SEAs that perform a larger set of tasks.
%

In practice, our methodology presents some challenges. The most important is the fabrication of elastic elements with arbitrary torque-elongation behavior. For instance, manufacturing the spring prescribed in Fig. \ref{fig:OptimalElasticElementConstrained} requires additional mechanical design as off-the-shelf springs normally have a linear torque-elongation relationship. However, manufacturing of nonlinear springs has been already explored in robotic applications \cite{Hawkes2018}. Jutte and Kota \cite{Jutte2008} introduced a generalized nonlinear spring synthesis methodology for any prescribed nonlinear load-displacement function. Custom springs can also be designed using topology optimization, especially for hyperelastic structures \cite{Chen2017}. Vanderborght et al. \cite{Vanderborght2013} summarized available techniques to produce nonlinear springs; cam, hypocycloid, and double-slider mechanisms coupled with linear springs are some examples of these techniques \cite{Realmuto2015, Thorson2011,Jung-JunPark2009}. It is important to note that some of these techniques may introduce energy losses, such as dissipated heat due to friction in the mechanisms, and result in elastic elements that may not be conservative. Materials with inherent nonlinear elasticity, such as polymers, may exhibit viscoelastic behavior with its respective energy losses \cite{Bolivar2016}. To account for these losses, our cost function can be extended to include viscous friction losses in the spring, and the optimization problem remains convex. However, we did not include these losses explicitly as elastic elements are very efficient in practice \cite{Hubicki2016}.

In summary, the proposed methodology designs conservative elastic elements that minimize energy consumption and peak power for arbitrary periodic reference trajectories subject to actuator constraints. Simultaneous minimization of energy consumption and peak power is not trivial and motivates the trade-off analysis described in Section \ref{sec:Simulation results}. The use of general conservative elastic elements may increase the range of operation in which the benefits of SEAs apply. Future work will focus on the implementation of the optimal elastic element and the formulation of a motor control algorithm to perform a task with minimum energy consumption that satisfies the torque-speed characteristics of the motor.

\bibliographystyle{asmems4}
\bibliography{JMR_17}


\begin{IEEEbiography}[{\includegraphics[width=1in,height=1.25in,clip,keepaspectratio]{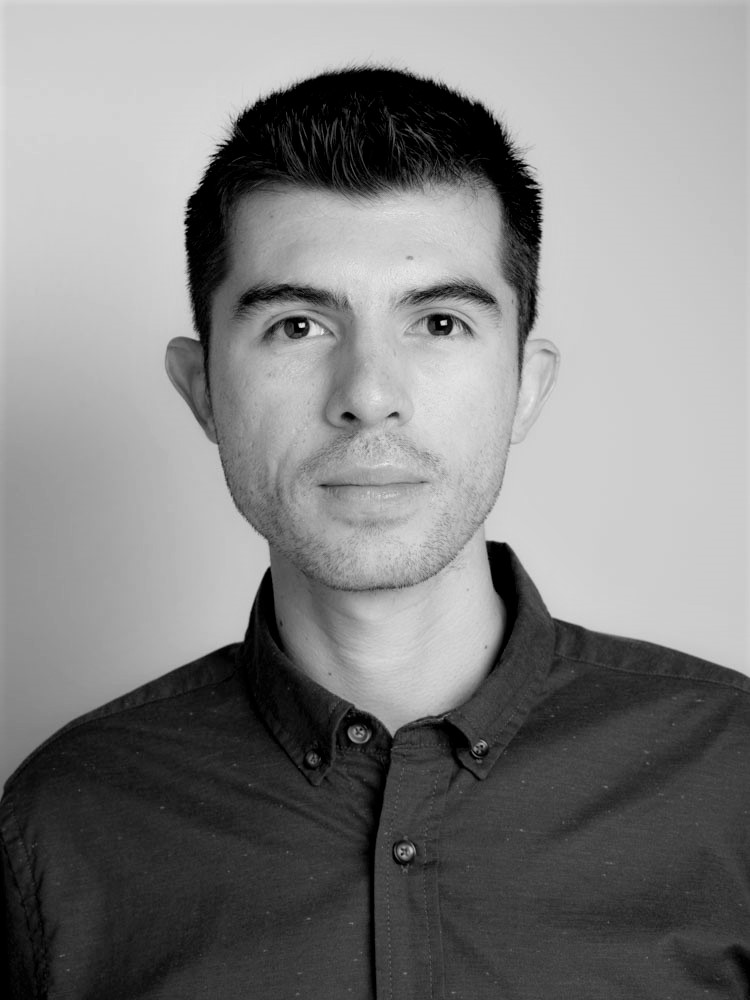}}]{Edgar Bol\'ivar}
(S’16) received the B.S. degree in Mechatronics Engineering from the Universidad Nacional de Colombia (2011). Currently, he is a Ph.D. student in Mechanical Engineering at the University of Texas at Dallas. He is with the Locomotor Control Systems Lab (LoCoLab) at the University of Texas at Dallas. His goal at the LoCoLab is to provide wearable robots with actuators and control systems that allow synergy between human and machine. Prior to joining the LoCoLab, he was undergraduate research assistant at the University of Wisconsin-Milwaukee and research engineer at the Universidad Nacional de Colombia.
\end{IEEEbiography}

\begin{IEEEbiography}[{\includegraphics[width=1in,height=1.25in,clip,keepaspectratio]{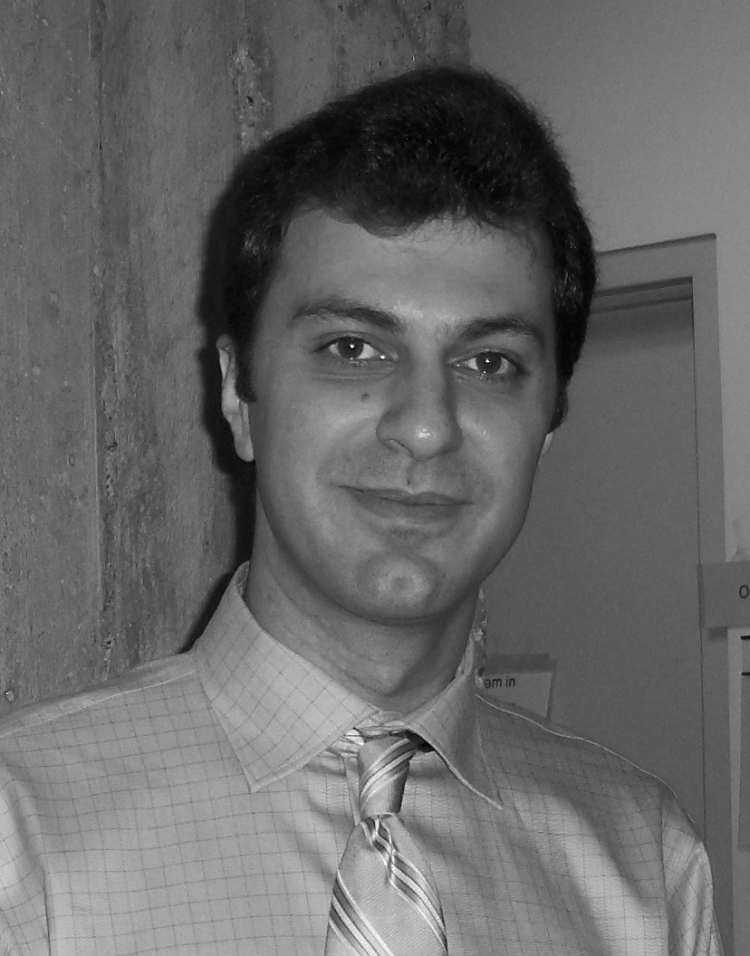}}]{Siavash Rezazadeh}
received his B.Sc. and M.Sc. from Sharif University of Technology, and his PhD from University of Alberta. Currently, he is with the Locomotor Control Systems Lab at University of Texas at Dallas, working on design and control prosthetic legs. Before joining Locolab, he worked at Dynamic Robotics Laboratory of Oregon State University on control of ATRIAS, a bipedal robot, for DARPA Robotics Challenge (DRC).
\end{IEEEbiography}

\begin{IEEEbiography}[{\includegraphics[width=1in,height=1.25in,clip,keepaspectratio]{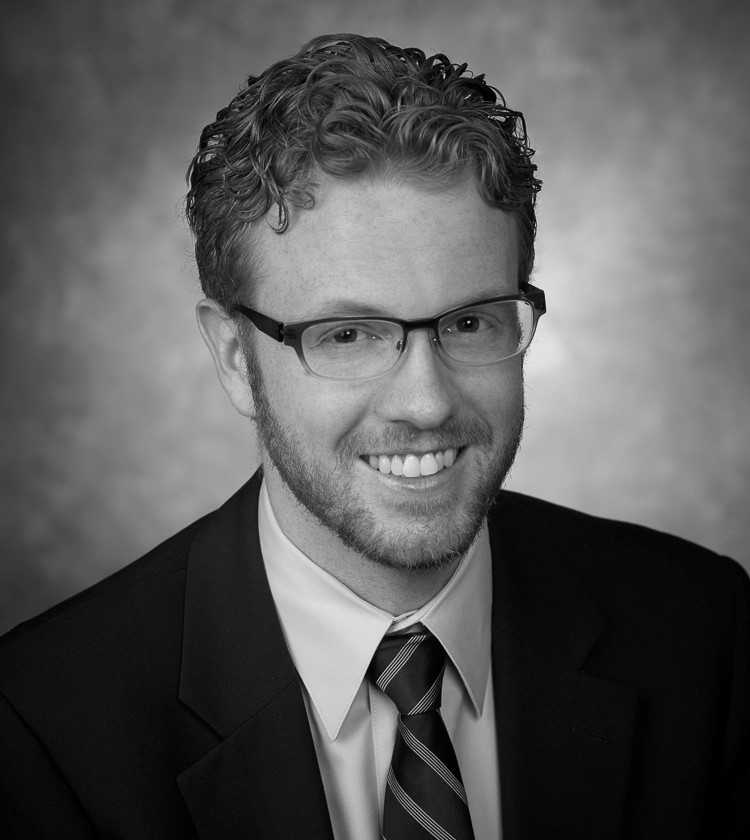}}]{Robert D. Gregg IV}
(S’08-M’10-SM’16) received the B.S. degree (2006) in electrical engineering and
computer sciences from the University of California, Berkeley and the M.S. (2007) and Ph.D. (2010) degrees in electrical and computer engineering from
the University of Illinois at Urbana-Champaign. He joined the Departments of Bioengineering and Mechanical Engineering at the University of Texas at Dallas as an Assistant Professor in June 2013. Prior to joining UT Dallas, he was a Research Scientist at the Rehabilitation Institute of Chicago and a Postdoctoral Fellow at Northwestern University. His research is in the control of bipedal locomotion with applications to autonomous and wearable robots.
\end{IEEEbiography}

\end{document}